\documentclass[sigconf]{aamas} 

\newif\ifmargincomments
\margincommentsfalse

\newif\ifextendedversion
\extendedversiontrue

\newif\iftasknetworks
\tasknetworksfalse

\usepackage{balance} %
\usepackage{graphicx}
\graphicspath{{.}{images/}}

\usepackage{color}
\usepackage[colorinlistoftodos,prependcaption,textsize=footnotesize]{todonotes}
\iftasknetworks
\ifextendedversion
\usepackage{minted}
\fi
\fi
\usepackage{algorithm}
\usepackage{algpseudocode}

\ifmargincomments

\newcommand{\frtodo}[1]{\todo[inline,color=green]{[fr] #1}}
\newcommand{\mmtodo}[1]{\todo[inline,color=red]{[mm] #1}}
\newcommand{\agtodo}[1]{\todo[inline,color=blue]{[ag] #1}}
\newcommand{\grtodo}[1]{\todo[inline,color=yellow]{[gr] #1}}
\newcommand{\revI}[1]{{\color{blue}#1}}
\else

\newcommand{\frtodo}[1]{}
\newcommand{\mmtodo}[1]{}
\newcommand{\agtodo}[1]{}
\newcommand{\grtodo}[1]{}
\newcommand{\revI}[1]{#1}
\fi

\newcommand{\gobble}[1]{}

\makeatletter
\gdef\@copyrightpermission{
 * denotes equal contribution.
 
  \begin{minipage}{0.2\columnwidth}
   \href{https://creativecommons.org/licenses/by/4.0/}{\includegraphics[width=0.90\textwidth]{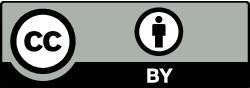}}
  \end{minipage}\hfill
  \begin{minipage}{0.8\columnwidth}
   \href{https://creativecommons.org/licenses/by/4.0/}{This work is licensed under a Creative Commons Attribution International 4.0 License.}
  \end{minipage}
  \vspace{1pt}
}
\makeatother

\setcopyright{ifaamas}
\acmConference[AAMAS '25]{Proc.\@ of the 24th International Conference
on Autonomous Agents and Multiagent Systems (AAMAS 2025)}{May 19 -- 23, 2025}
{Detroit, Michigan, USA}{Y.~Vorobeychik, S.~Das, A.~Nowé  (eds.)}
\copyrightyear{2025}
\acmYear{2025}
\acmDOI{}
\acmPrice{}
\acmISBN{}

\acmSubmissionID{580}

\title[Planning and scheduling on the lunar CADRE mission]{Planning, scheduling, and execution on the Moon: the CADRE technology demonstration mission}

\author{Gregg Rabideau*}
\affiliation{
  \institution{Jet Propulsion Laboratory - California Institute of Technology}
  \city{Pasadena, CA}
  \country{USA}}
\email{gregg.rabideau@jpl.nasa.gov}

\author{Joseph Russino*}
\affiliation{
  \institution{Jet Propulsion Laboratory - California Institute of Technology}
  \city{Pasadena, CA}
  \country{USA}}
\email{joseph.a.russino@jpl.nasa.gov}

\author{Andrew Branch}
\affiliation{
  \institution{Jet Propulsion Laboratory - California Institute of Technology}
  \city{Pasadena, CA}
  \country{USA}}
\email{andrew.branch@jpl.nasa.gov}

\author{Nihal Dhamani}
\affiliation{
  \institution{Jet Propulsion Laboratory - California Institute of Technology}
  \city{Pasadena, CA}
  \country{USA}}
\email{nihal.n.dhamani@jpl.nasa.gov}

\author{Tiago Stegun Vaquero}
\affiliation{
  \institution{Jet Propulsion Laboratory - California Institute of Technology}
  \city{Pasadena, CA}
  \country{USA}}
\email{tiago.stegun.vaquero@jpl.nasa.gov}

\author{Steve Chien}
\affiliation{
  \institution{Jet Propulsion Laboratory - California Institute of Technology}
  \city{Pasadena, CA}
  \country{USA}}
\email{steve.a.chien@jpl.nasa.gov}

\author{Jean-Pierre de la Croix}
\affiliation{
  \institution{Jet Propulsion Laboratory - California Institute of Technology}
  \city{Pasadena, CA}
  \country{USA}}
\email{jean-pierre.de.la.croix@jpl.nasa.gov}

\author{Federico Rossi}
\affiliation{
  \institution{Jet Propulsion Laboratory - California Institute of Technology}
  \city{Pasadena, CA}
  \country{USA}}
\email{federico.rossi@jpl.nasa.gov}

\begin{abstract}
NASA's Cooperative Autonomous Distributed Robotic Exploration (CADRE) mission, slated for flight to the Moon's Reiner Gamma region in \revI{2025/2026}, is designed to demonstrate multi-agent autonomous exploration of the Lunar surface and sub-surface.
A team of three robots and a base station will autonomously explore a region near the lander, collecting the data required for 3D reconstruction of the surface with no human input; and then autonomously perform distributed sensing with multi-static ground penetrating radars (GPR), driving in formation while performing coordinated radar soundings to create a map of the subsurface.
At the core of CADRE's software architecture is a novel autonomous, distributed planning, scheduling, and execution (PS\&E) system. The system coordinates the robots' activities, planning and executing tasks that require multiple robots' participation while ensuring that each individual robot's thermal and power resources stay within prescribed bounds, and respecting ground-prescribed sleep-wake cycles. The system uses a centralized-planning, distributed-execution paradigm, and a leader election mechanism ensures robustness to failures of individual agents.
In this paper, we describe the architecture of CADRE's \pse system; discuss its design rationale; and report on verification and validation (V\&V) testing of the system on CADRE's hardware in preparation for deployment on the Moon. 

\end{abstract}


\newcommand{\BibTeX}{\rm B\kern-.05em{\sc i\kern-.025em b}\kern-.08em\TeX}
\newcommand{\pse}{PS\&E~}

\begin{document}

\pagestyle{fancy}
\fancyhead{}

\maketitle

\section{Introduction}
Distributed instruments hold great promise to unlock key questions in planetary science that are inaccessible to single-point measurements. The ability to collect time-synchronized and cross-calibrated measurements from spatially distributed\gobble{, and potentially moving,} locations is critical to investigations of subjects as diverse as atmospheric circulation on Mars, Venus, and Titan \cite{haberle2014preliminary}; sub-surface compositions of rocky and icy moons \cite{netsag2010,Vance}; and  seismic activity on Venus \cite{RossiSaboiaEtAl2023}.

The traditional operations paradigm for space exploration missions relies on on-board execution of sequences, developed by human operators on the ground, that specify which activities should be performed at which time by the spacecraft or rover \cite{sellmaier2022spacecraft}; however, such an approach generally does not scale to multi-agent systems, which must cope with bandwidth and latency constraints between individual \revI{agents}; heterogeneous availability of resources (e.g., thermal and power conditions that may vary among agents); and potentially time-varying participation, as some agents may become temporarily, or permanently, unavailable.

These considerations motivate the development of multi-agent autonomous planning, scheduling, and execution (PS\&E) tools for future space exploration missions. Many approaches for multi-agent \pse have been proposed in the literature; however, to date, no approach has been developed and tested to a level of \gobble{maturity (and, specifically,} technology readiness \cite{mankins1995technology}\gobble{)} sufficient for infusion in future spaceflight missions.
 
 \begin{figure}[h]
 \includegraphics[width=\columnwidth]{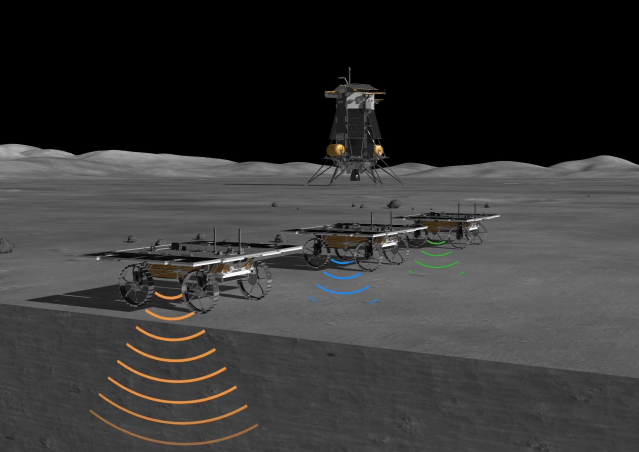}
 \caption{Overview of the CADRE mission. \gobble{Three autonomous rovers and a lander-mounted base station will cooperatively perform autonomous exploration and collect multi-static ground penetrating radar measurements of the Moon's Reiner Gamma region.}}
 \label{fig:cadre-hero}
 \vspace{-2.5em}
 \end{figure}
 
NASA's Cooperative Autonomous Distributed Robotic Exploration (CADRE) mission, shown in Figure \ref{fig:cadre-hero}, aims to fill this gap. CADRE will fly a team of four agents, including three mobile rovers and a static base station, to the Moon's Reiner Gamma region in \revI{2025/2026}. The goal of CADRE is to demonstrate the high technology readiness of cooperative autonomy, that is, the ability for a team of autonomous agents to elaborate high-level commands from ground (e.g., ``explore this region'') into commands for the rovers' mobility system and on-board instruments. \revI{Specifically, the rovers will cooperatively explore a prescribed region and collect multi-static ground penetrating radar measurements; the base station will act as a communication gateway between the rovers and Earth, and aid the team by performing computational tasks.}
 
CADRE's on-board autonomy encompasses autonomous planning, scheduling, and execution; multi-agent motion-planning;  frontier exploration; and single-agent mobility, including localization, mapping, and single-agent motion planning. In this paper, we focus on the autonomous \pse architecture, i.e., on the module that elaborates high-level commands into tasks executed by individual vehicles. We refer the interested reader to \cite{DeLaCroixRossiEa2024} for a discussion of CADRE's overall autonomy architecture.

\subsection{State of the Art}

\paragraph{Multi-agent planning, scheduling, and execution} A vast number of approaches have been proposed for multi-agent planning, scheduling, and execution in the academic community. Coordination approaches include \emph{centralized planning}, where the multi-agent system is effectively treated as a single agent; \emph{leader-election}-based approaches, in which an agent is elected as leader and performs centralized planning for the entire team; \emph{explicit-coordination} approaches (including auctions and broadcast-decentralized algorithms), where agents either bid on tasks to be completed based on their state and resources (e.g., \cite{choi2009consensus}), or explicitly broadcast contention information (e.g., \cite{parjan-jais2023-mas,zilberstein-icaps-2024}); \emph{shared-world} (or \emph{implicit coordination}) approaches, in which each agent maintains a world model through agent-to-agent information sharing, performs planning for the entire system based on the world model, and executes its part of the plan (e.g., \cite{wolf2017caracas,adams2024distributed});  and \emph{emerging behavior} approaches, in which agents communicate with their immediate neighbors and select the next task based on simple heuristics (e.g., \cite{Werfel2014termite}).

We refer the reader to \cite{RossiBandyopadhyayEtAl2021} for a survey of algorithmic approaches to multi-agent decision-making, including planning and scheduling.

\paragraph{Planning and scheduling in space} The majority of the approaches outlined above approaches have not been demonstrated at a technology readiness level (TRL) sufficient for infusion in future missions.

A notable exception is the Distributed Spacecraft Autonomy (DSA) mission \cite{cellucci2020distributed,adams2023overview,adams2024distributed} on NASA's Starling mission. DSA has demonstrated shared-world planning of scientific observations in Earth orbit, where all agents build a common world model, and the planning problem is solved as an integer linear program. 

However, a key difference makes DSA's approach not immediately applicable to CADRE's concept of operations. 
DSA operates in low Earth orbit, where performance of wireless inter-satellite communication links is comparatively predictable; this makes a shared-world approach appropriate, since it is highly likely that, with proper design of the telecommunication system, all agents will converge to the same world model. In contrast, CADRE's domain of operations is the Lunar surface, where surface obstructions and the behavior of disturbed regolith may strongly affect inter-agent communications; this makes a shared-world approach less desirable, since communication failures can result in complex and hard-to-diagnose miscoordination between  agents. 

Several approaches to \emph{single-agent} planning and scheduling have been demonstrated in space. A notable example is the Perseverance Mars rover's Onboard Planner \cite{verma2023autonomous}, in operational use since 2023\gobble{, which schedules and executes tasks on the rover in response to the rover's thermal and power state and to environmental conditions}.
The planner holds promise to significantly increase the rover's mission productivity compared to current ground-in-the-loop approaches (which see the rover idle for up to 28\% of the time \cite{gaines-doran-justice-et-al-2016,gaines-doran-justice-et-al-IWPSS-2017}), and also reduce rover energy usage by over 10\% for a given campaign, saving resources for more opportunistic science investigations.  
A second notable example is CASPER, flown on NASA's EO-1 spacecraft \cite{chien2005eo1}. On-board event detection  from hyperspectral imagery was used to detect thermal anomalies (e.g., lava flows), clouds, floods, and other features; CASPER then replanned observations in response to these detections\gobble{, e.g., rescheduling observations obstructed by high cloud coverage}.
A key commonality between all of these approaches\gobble{, including DSA,} is the strong decoupling between information-sharing and planning --- that is, the assumption that either the planner has access to full system information (in Onboard Planner and CASPER), or that every planner has access to the same information (as in DSA). This assumption is natural for single-agent systems and reasonable for orbiting distributed systems; however, it is significantly more restrictive for surface systems.\gobble{, where line-of-sight and Fresnel zone obstructions may degrade communications between agents in hard-to-predict ways.}

\subsection{Contribution}

Our contribution is the development and testing of a leader-election-based multi-agent planning, scheduling, and execution architecture designed for operations on planetary surfaces. %
The proposed approach explicitly reasons about the information required for planning and for execution, and does \emph{not} assume that all agents have access to the same information;
careful, coupled design of the planning architecture and the task definitions results in reduced inter-agent communications compared to shared-world and auction-based approaches, ensuring good performance as well as resilience to disrupted inter-agent communications.

The rest of this paper is structured as follows. In Section \ref{sec:architecture}, we describe CADRE's \pse architecture. Section \ref{sec:tasknets} describes how the proposed task definitions implement CADRE's concept of operations, and discusses how tasks are designed to use local information as much as practical to reduce communication. Section \ref{sec:testing} discusses the testing, verification, and validation of the proposed architecture and task definitions. Finally, we draw our conclusions in Section \ref{sec:conclusion}.

\section{Planning, scheduling, and execution architecture}
\label{sec:architecture}

\begin{figure}[h]
\centering
\includegraphics[width=\columnwidth]{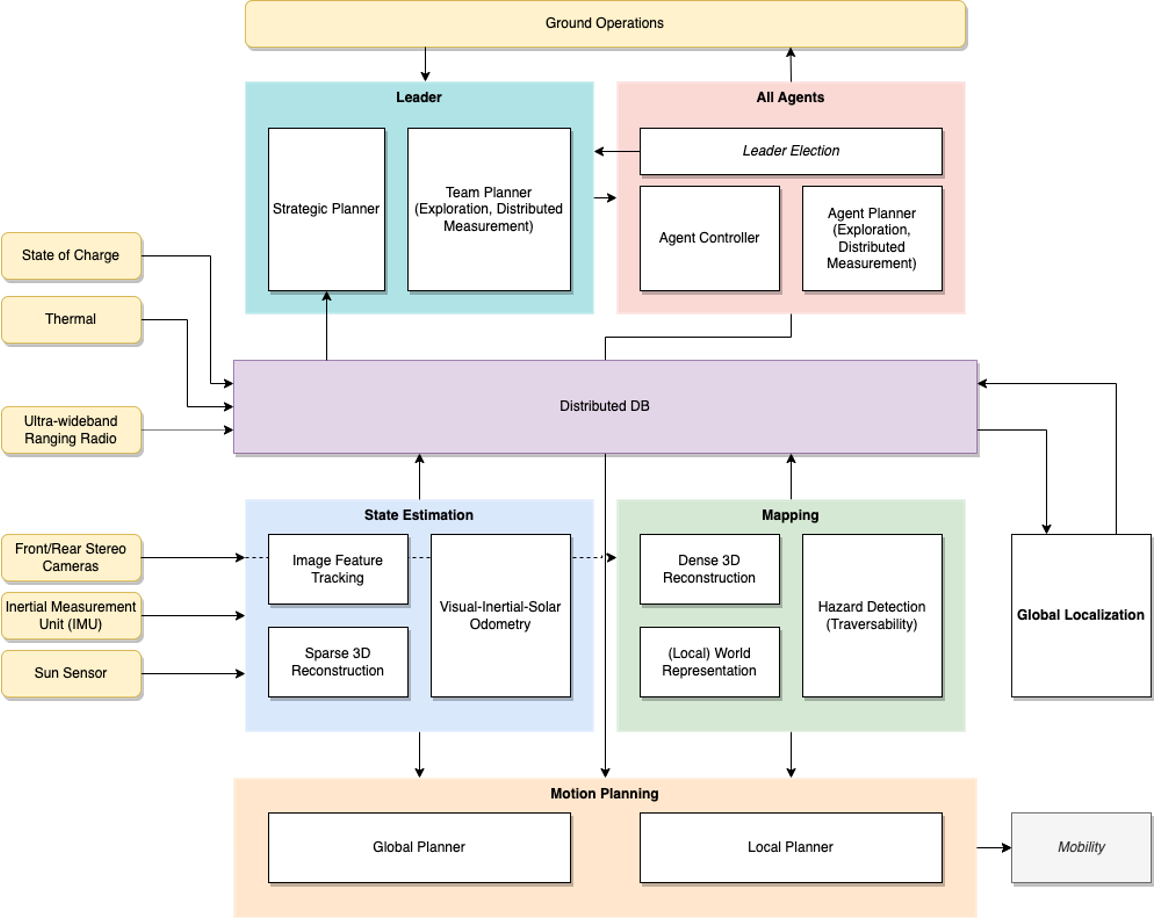}
\caption{Architecture of CADRE's flight software. The second and third row of blocks from the top represent the \pse architecture. \gobble{\revI{Communication between modules, commanding, and telemetry are implemented using the F' flight software framework \cite{bocchino2018fprime}.}}}
\label{fig:architecture:architecture}
\end{figure}

CADRE's \pse architecture, shown in Figure \ref{fig:architecture:architecture}, relies on four sets of modules:
\begin{itemize}
\item a \emph{leader election} module that selects one agent to perform planning for the entire team;
\item a lightweight \emph{shared state database} that synchronizes selected state information from all agents to the elected leader;
\item a \emph{strategic planning} module that plans and schedules tasks for all agents; and
\item a set of \emph{agent controllers}, one per agent, that execute the scheduled tasks; monitor task progress and constraint satisfaction; and inform the strategic planning module of task success and failure and of constraint violations that affect other agents' tasks.
\end{itemize}

\subsection{Leader Election}

A leader election module ensures that every agent in the team agrees on the designation of a single agent as leader at any given time. A distributed minimum spanning tree algorithm, GHS \cite{gallager1983distributed}, is used to select a unique appointer among the agents every ten seconds; the appointer then selects as leader the agent whose state is farthest from violating thermal and power constraints, and broadcasts the leader designation to all agents. A hysteresis mechanism is used to ensure that the leader is not changed unless large changes in its state are observed. The leader election module also \revI{periodically} chooses a ``designated survivor'', an agent tasked with holding a copy of the leader's information, making recovery from loss of a leader faster. \revI{The architecture ensures that loss of the leader results in at most ten seconds of idle time before planning recommences on a new leader}. We refer the reader to \cite{AlbeeBhamidipatiEa2024} for a detailed description of the leader election approach and implementation.

\subsection{Shared State Database}

A shared state database (SSDB) allows agents to synchronize key state information periodically. The shared state database is based on SQLite, and synchronization is implemented through a custom application layer. The implementation deliberately does not guarantee consistency between the agents in the sense of the CAP theorem \cite{gilbert2002brewer} (i.e., different agents' databases may contain different information at any given time), prioritizing availability over consistency and providing robustness to network partitions. Critically, due to the elected-leader architecture, such lack of consistency will \emph{not} result in uncoordinated action among the agents, although it may result in suboptimal behavior if the leader has stale information. State information that (i) is required for strategic planning and (ii) cannot be inferred locally is synchronized by all agents to the leader \revI{and the designated survivor} periodically, every 10 seconds. \revI{All information required for planning is stored on at least two agents, ensuring robustness to loss of a single agent; information is only lost if both the leader and the designated survivor fail before a new leader and survivor can be elected (which happens within 10s), and before information can be synchronized to them}.  We refer the reader to \cite{SaboiaRossiEa2024} for a detailed description of CADRE's SSDB.

\subsection{Centralized planning}
\label{sec:architecture:planner}

A strategic planning module runs on the leader and plans and schedules tasks for all agents.
The planner module continually evaluates the plan at a set cadence (currently, 1Hz) to evaluate re-planning triggers, commit new tasks, delete old tasks, and check multi-agent constraints. Triggers for re-planning include conflict detection, task failure, as well as at specific milestones in the experiment (e.g., when an exploration or distributed measurement task is completed). When a task's scheduled start time falls within a certain window (currently set to 5 seconds), that task is handed off, or \emph{committed}, to the relevant agent controller, described next. 
\revI{Since the planner has visibility into the state of all agents, planning of tasks that require cooperation between multiple agents is straightforward. Enforcement of coordination during decentralized execution is more complex, and is discussed next in Section \ref{sec:architecture:abort}.}
The planner reasons explicitly about agent availability, which is communicated to it directly from the agent controller modules, in order to avoid scheduling and committing tasks for an agent that is not actively running a controller module. The strategic planner and agent controller modules for CADRE are implemented using Multi-mission EXECutive (MEXEC)~\cite{troesch_mexec_asteria_intex2020}, a resource-aware onboard planning and execution software that uses task networks to generate and execute conflict-free schedules. 
In task networks, tasks are represented as a tuple containing a flight software (FSW) command to execute the task; an expected duration; a set of constraints; and a set of expected impacts.
Constraints include both state constraints, which require that system states remain between prescribed maximum and minimum values; and precedence constraints, where one task must be completed before another can be started. State constraints include both pre-execution constraints that must be satisfied before a task can be started, and maintenance constraints that must be satisfied throughout execution.
Impacts represent the expected change in state from executing the given task (e.g., the expected power draw from driving). During planning, a task's impacts are used to predict the future evolution of the system state; constraints for subsequently-scheduled tasks are then verified against these predicted values. We refer the reader to \cite{troesch_iwpss2019_robustmapping,troesch_mexec_asteria_intex2020} for a detailed description of the task network representation. A priority-based insertion heuristic is used for planning, as detailed in Section \ref{sec:tasknets:encoding:algorithm}. %

\subsection{Decentralized execution}
\label{sec:architecture:controller}

Each agent runs an agent controller module, which is responsible for managing task execution. \gobble{With the exception of the instance running on the leader agent, each} Each agent controller is only able to observe its own agent's state. The agent controller checks task constraints before and during execution; it has the authority to delay the start of a task if any of its starting constraints is unmet, and also to declare a task as ``failed'' if any of its constraints is violated during the course of execution. In the latter case, the controller issues cleanup commands needed to keep the agent in a recoverable state. The agent controller reports the execution state of all tasks under its authority back to the strategic planner \revI{on the leader}, which uses this information \revI{to trigger} re-scheduling. %

\subsection{Multi-agent constraint checking}
\label{sec:architecture:abort}
Inter-agent coordination of tasks is maintained during execution through the \revI{enforcement} of multi-agent constraints, \revI{performed on the leader}. The information available to individual agents is limited to the state of that agent, with the exception of the leader. As such, each agent does not have the necessary information to evaluate all task constraints for tasks \revI{requiring coordination among multiple agents, where a constraint violation on another agent's state may trigger a stop}. Constraints that cannot be evaluated on \revI{individual} agents due to lack of information are referred to as ``multi-agent constraints''. These constraints are evaluated on the leader. %
Similar to regular constraints, multi-agent constraints include both pre-execution constraints and maintenance constraints. \revI{For multi-agent constraints}, however, a pre-execution constraint cannot \revI{directly} delay the start of a task, since the \revI{leader} is not controlling the task. Instead, the \revI{leader} does a one-time check to verify that the constraint was satisfied within a short time after the task start time. Maintenance constraints are 
continuously checked throughout the execution of a task.
For any constraint found to be violated, an ``abort task'' message is sent from the leader to the agent executing the corresponding task. On receiving the ``abort task'' message, the agent's controller fails the task, causing the leader to replan.

\subsection{Integration with leader election}

Only the leader agent performs planning and scheduling. However, the leader may change at any time. For this reason, an instance of the strategic planner module \revI{exists} on every agent; on the leader, the strategic planner is in an active state and schedules and commit tasks, whereas on the non-leader agents the strategic planner module is in an inactive state. To achieve an orderly transition from one leader to the next, (i) only one strategic planner may be actively scheduling and committing tasks at a given moment, and (ii) 
when a new leader is elected, no agent controller may be actively commanding or tracking tasks that were scheduled and committed by a previous leader.

To accomplish (i), the strategic planner \gobble{module} listens to leader election updates and also to acknowledgement messages from each agent controller\gobble{ module}. When an active strategic planner receives an update from leader election indicating that its agent is no longer the leader, it commits no new tasks and immediately transitions to the inactive state. When an inactive strategic planner receives an update indicating that its agent is now the leader, it waits for acknowledgement messages from each participating agent controller, and only transitions to the active state once all agents have acknowledged that it is leader and that they are ready to accept tasks.

To accomplish (ii), the agent controller modules also listen to leader election updates. As soon as an agent controller receives an update from leader election indicating that the leader has changed, it \ purges the set of tasks that are under its authority. Any task that has not yet been commanded is immediately dropped. Cleanup commands are issued for any task that has them, and then the controller waits for all \gobble{of the} running tasks \gobble{that it is tracking} to complete. Only once all tasks under its authority are complete does the agent controller send acknowledgement to the strategic planner on the new leader that it is ready to receive tasks from it.

This approach\gobble{ serves to minimize state that is persisted when the leader changes, and} allows the system to be resilient to unexpected loss of a leader, as the new leader does not rely on information shared from the previous one.

\subsection{Remarks on inter-agent communication}

We note that the \pse architecture design assumes that a low-bandwidth, potentially high-latency, bidirectional communication channel is continuously available between agents and the elected leader. If the communication channel between agents may drop arbitrary packets (and potentially all packets), no coordination can be achieved in general --- a problem known as the \emph{coordinated attack} problem \cite{Gray1978}. %
We strive to minimize the amount of information sent over links to maximize the likelihood of such information successfully being exchanged, and we design the system to be robust to latency in message delivery. %
\revI{Specifically,} if state information from an agent does not propagate to the leader in a timely manner, the leader's plan will rely on stale information from that agent, which results in \gobble{sub-optimal and} potentially infeasible plans; however, if the resulting plan indeed violates the agent's constraints, the agent controller will refuse to execute the task based on its local information, ensuring safety.
Similarly, if a committed task fails to propagate in a timely manner from the leader to the agent that should execute it, the agent will re-evaluate whether the task remains feasible upon receiving it; if the task is still feasible, it will execute it anyway, and if it is no longer feasible, it will declare it as failed.

For tasks that rely on coordination between agents, task failure on one agent requires stopping all other agents and re-planning; this is accomplished by communicating the task failure from that agent to the leader, triggering a multi-agent constraint failure, which in turn causes the leader to send ``abort'' messages to other affected agents. If such messages are delayed, agents may act in an uncoordinated manner for some time. We acknowledge this as a limitation of the proposed approach, and conjecture that it may be a fundamental limitation in presence of communication latency; to minimize the likelihood of occurrence, we reduce the size of ``abort'' messages to a single task ID, in order to empirically maximize the likelihood of timely delivery even on a disrupted communication link.

\section{Planning CADRE's mission}
\label{sec:tasknets}

\subsection{Concept of Operations}
\label{sec:conops}

CADRE will perform two experiments to demonstrate multi-agent autonomy: exploration and distributed sensing with multi-static radar.
An FPGA, not controlled by autonomy, controls a timer that shuts down all agents at 25 and 55 minutes past the hour, and restarts them 0 and 60 minutes past the hour  \cite{DeLaCroixRossiEa2024}. Within each 25-minute cycle, the team must make progress towards the active experiment as follows.

\subsubsection{Exploration}

\begin{figure}[h]
\includegraphics[width=.42\columnwidth]{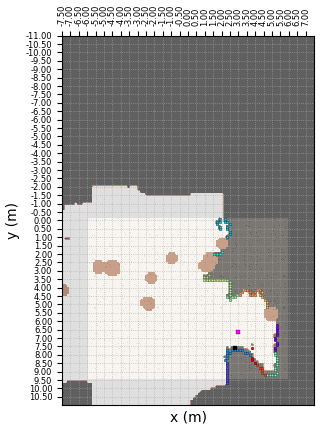}
\includegraphics[width=.42\columnwidth]{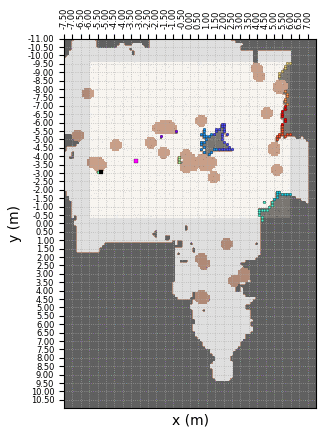}
\vspace{-.75em}
\caption{Exploration: two rovers (magenta dots) explore the sub-region assigned to them (lighter rectangle) by selecting a frontier (\gobble{denoted by the }colorful segments) between traversable (off-white) and unexplored (light gray) portions of the sub-region.}
\label{fig:conops:exploration}
\vspace{-1em}
\end{figure}

During the exploration experiment, ground operators uplink the boundary of a region to explore\gobble{ (represented as a polygon)} to the team. The content of the region is initially unknown to the agents. The goal of the experiment is for the agents to collectively observe every portion of the region that is reachable from their initial locations with on-board stereo cameras; and to classify every reachable portion of the region as ``traversable'' or ``obstacle''. To do this, the CADRE team partitions unknown portions of the region into as many sub-regions as there are rovers, and assigns each sub-region to a rover; each rover then explores its own sub-region using frontier-based exploration \cite{yamauchi1997frontier}, as shown in Figure \ref{fig:conops:exploration}. Maps are periodically synchronized to the leader, which merges them in a joint map, and sub-regions are periodically re-computed as the unknown portions of the map shrink.
We refer the reader to \cite{nayak2024exploration} for a detailed description of the exploration architecture.

\subsubsection{Distributed Sensing}

\begin{figure}[h]
\includegraphics[width=.67\columnwidth]{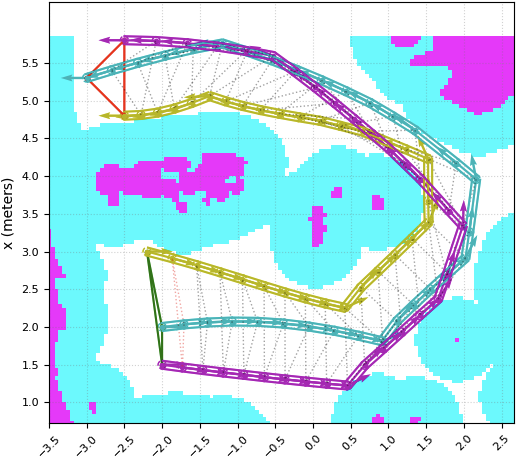}
\vspace{-.75em}
\caption{Distributed sensing\gobble{ plans a joint in-formation trajectory for all rovers}: each rover is assigned a ``tube'' around its nominal trajectory, and must travel inside the tube while satisfying temporal constraints.}
\label{fig:conops:formation}
\end{figure}

During the distributed sensing experiment, rovers are tasked with driving in formation between assigned waypoints while collecting multi-static ground-penetrating radar readings. Ground operators specify a waypoint to reach as a team, typically tens of meters away from the rovers' initial location; a formation to maintain, specified as a set of inter-rover distances; and a maximum allowable deviation from the prescribed formation that will, if not violated, ensure sufficient quality of the radar data. Agents must plan a set of trajectories that will reach the prescribed waypoint while staying in formation. The map of the region where the experiment is performed may be partially unknown to the rovers (that is, the distributed sensing experiment may stray outside the exploration region area). Thus, the rovers are likely to encounter initially-unknown obstacles that require replanning. Team-level planning is performed by the leader through sampling-based motion planning \cite{karaman2011sampling} on the rovers' joint state space; individual rovers are then assigned time-stamped corridors around the computed team trajectory (shown in Figure \ref{fig:conops:formation}). If every rover remains within its own time-stamped corridor, the maximum allowable deviation between rover distances is guaranteed not to be violated. If a rover is unable to remain within the allocated corridor, all other rovers must stop; the leader then computes a new collision-free team trajectory.

\subsection{Task Network Encoding}
\label{sec:tasknets:encoding}

\subsubsection{Tasks} To encode the exploration and distributed sensing behaviors, we build upon single-agent localization, mapping, and mobility tasks to create higher-level behaviors that allow the system to operated autonomously as a whole. PS\&E's  \gobble{key} role is to \gobble{track and }manage the interaction between these tasks, and their system-level impacts (e.g., their power consumption and impact on rover temperature). To accomplish this, we model a set of abstract tasks sufficient to predict system impacts and constrain agent behavior to safe and desirable actions.\gobble{ The strategic planner uses these abstract tasks to schedule as much activity as possible within these constraints.}

Specifically, the tasks considered in \pse are: 
\begin{itemize}
    \item \emph{SSDB map synchronization}: every \revI{rover} copies its new maps and pose to the leader; this task must be executed prior to exploration and formation navigation planning to ensure that the leader has access to up-to-date maps from all rovers;
    \item \emph{SSDB backup}: the designated survivor maintains a backup of \revI{all agents'} SSDB to \revI{insure against data loss and} enable efficient leader reelection\gobble{, if necessary}; %
    \item \emph{exploration navigation planning}: the region to explore is divided in as many subregions as there are rovers; the task is executed on the leader during the exploration experiment;
    \item \emph{driving to explore}: each rover is tasked to explore its sub-region via frontier-based exploration; the task is executed on \revI{each} participating rover during the exploration experiment;
    \item \emph{formation navigation planning}: a set of trajectory ``tubes'' is computed \revI{for the participating rovers}; the task is executed on the leader during the distributed sensing experiment; 
    \item \emph{driving in formation}: each rover is tasked with following the tube around its trajectory; the task is executed on \revI{participating} rovers during the distributed sensing experiment;
    \item \emph{stopping a drive early}:  while driving tasks stop on their own when \revI{they reach their destination}, this task is used to \gobble{explicitly} stop early when time or resource constraints are close to violation;
    \item \emph{shutting down the agent software and electronics}: if time and resources continue to be insufficient, the task puts the agent in a low power mode.
\end{itemize}

\begin{figure}[h]
    \centering
    \includegraphics[width=0.9\linewidth]{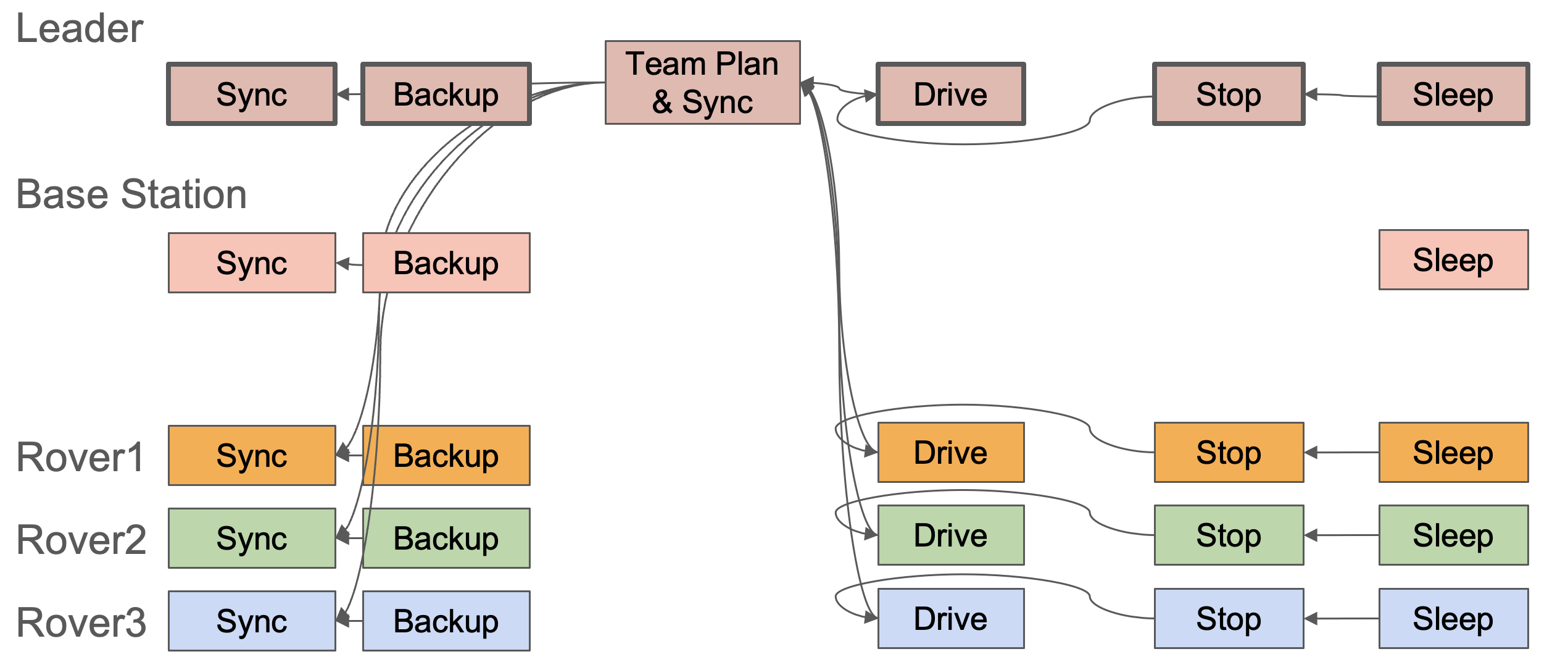}
    \vspace{-.5em}
  \caption{Abstract representation of the task network for both formation sensing and exploration. Arrows represent precedence constraints\gobble{; one task can only be scheduled if the preceding task has been successfully scheduled}. Resource and multi-agent constraints are not shown.}
    \label{fig:tasknet-formation-sensing}
\end{figure}

Figure \ref{fig:tasknet-formation-sensing} shows the set of tasks considered for the formation driving experiment, and selected inter-task constraints.

Each task (and in particular driving) involves a complex set of individual behaviors, which are modeled with a combined sets of constraints and impacts.
The energy consumed and heat produced by the CPU during image processing, for example, is rolled up with the power and heat from driving motors. 
Closed-loop replanning is used to correct for inaccuracies in the impact models. Constraints are used for both ensuring the proper task ordering as well as preventing over-subscription of system resources.

\subsubsection{Resources} The three modeled system resources are:
\begin{itemize}
    \item time: tasks must be performed within the maximum allotted 25 minute interval;
    \item battery state-of-charge (SOC): \revI{rover} batteries are recharged by solar arrays, but drained while performing tasks; \revI{the base station is powered by the lander, and has no battery};
    \item CPU temperature: operating for long durations close to Lunar noon can exceed the maximum operating temperature.
\end{itemize}

The strategic planner must anticipate the FPGA-commanded shutdown at 25 and 55 minutes past the hour; stop driving; and put the agent in a low-power mode before this timer expires. This allows logs and other data to be saved before power off. In addition, the battery SOC must be maintained about a minimum \revI{threshold} (e.g. 20\%). Simulation and analysis has shown that, during periods with the sun high overhead, stopping the drive and other tasks is sufficient to allow the solar panels to recharge the batteries. At other times, the agents must be put into a low-power mode through the shutdown task.
Temperature, however, limits agent operations at exactly the opposite periods of the lunar day. When the sun is low in the sky, lunar temperatures are low enough that there is no risk of overheating the CPU\gobble{, even at full operation}. But around lunar noon, the rovers can only drive for a few minutes before the CPU temperature exceeds its operational limit of 65$^\circ$ C. \gobble{In addition to resources, several logical states of the rover are tracked to ensure that tasks are scheduled only for those rovers that are participating in team activities, awake, and ready to perform those activities; these states are discussed next in Section \ref{sec:tasknets:multiagent}.} %

\subsubsection{Resource Models} Both battery SOC and temperature are measured through onboard sensors; whenever replanning is performed, predicted impact for future tasks are applied starting from this current state. Because of this, and the comparatively short planning horizon, impacts use simple models of resource change. 

For battery SOC, we use a roll-up of the power loads from all behaviors started by the task. Using this, and the battery capacity, we get a \gobble{simple,} linear prediction of the battery SOC. For heating discharge, and re-charging provided by the solar arrays, we use a coarse estimate based on the lunar time-of-day. The power required by heaters impacts a background discharging rate, which is reduced during power-consuming tasks to account for supplemental heating from power loads of that task (assuming that a uniform 80\% of the load will be dissipated as usable heat). Degradation in solar array performance due to rover tilt is omitted for simplicity.

Temperature proved more difficult to model.
The impact of tasks on rover temperature depends in a nonlinear way on the vehicle's starting temperature, the ambient temperature, rover tilt, and the power loads being applied. Simulations were performed at solar angles between $20^\circ$ and $90^\circ$ with $10^\circ$ steps; each simulation cycled through the different rover operating modes (i.e., the set of tasks being executed) multiple times. These simulations helped identify which factors have the largest effect on temperature rate-of-change. Sun angle and rover operating mode were identified as the two key factors affecting temperature rate of change. We also identified the CPU temperature as the most constrained temperature resource in all simulations; accordingly, we only model CPU temperature in planning, effectively treating each rover as a one-node thermal model. %
Surprisingly, we found that the CPU temperature at the start of an operating mode had a minimal impact on the heating rate; therefore, we omitted it from the planning model.
We did observe that the rate of change of temperature was markedly nonlinear, with an initial steep change that leveled off after about two minutes --- in line with the expected physical behavior of heat conduction and radiation.
The existing task model in MEXEC, however, only supports linear rates. We considered implementing a more accurate model that divides tasks into multiple parts to better capture this behavior. However, simulations and analysis showed that the added complexity would provide a relatively small benefit compared to more frequent replanning; accordingly, a linear model was used.

\subsubsection{Planning and Scheduling Algorithm} %
\label{sec:tasknets:encoding:algorithm}
When the system replans, it always has one of two goals, namely, to continue exploring or performing distributed measurements. The tasks performed during the previous wake cycle are not explicitly part of the state considered for planning: their impacts are fully captured by the rovers' maps for exploration, and the rovers' locations for distributed measurement. 
This allows us to use one simple scheduling process, regardless of when and why the system performs re-scheduling:
\begin{itemize}
    \item un-schedule any task not committed for execution;
    \item read and apply the current system state;
    \item \revI{greedily} schedule un-executed tasks in priority order, \revI{scheduling each task as close as possible to its preferred start time while ensuring constraint satisfaction, and} rejecting any task that cannot be scheduled without constraint violations.
\end{itemize}

\revI{Pseudocode for the planning algorithm is reported in \ifextendedversion Appendix \ref{apx:simpleplanner}. \else the Extended Version \cite{RabideauRussinoEaAAMAS24EV}. \fi}
The choice of this simple scheduling process has implications on how the task network is defined. First, priority order scheduling meant that we cannot assume that lower priority tasks will exist when a higher priority task is scheduled. Second, because no additional tasks are created during planning, the  task network must contain all of the tasks we will ever need. For example, because driving requires a team plan to be generated, the team planning task must be higher priority, while the driving task must be constrained to prevent scheduling if a team planning task does not exist. Moreover, because multiple driving cycles could be scheduled, the task network includes multiple instances of all tasks. But a second instance of driving should not be scheduled if one already exists, unless that first drive was stopped or completed. Constraints were used to ensure proper task ordering and allow multiple exploration and distributed measurement cycles.

\subsection{Multi-agent Task Networks}
\label{sec:tasknets:multiagent}

For tasks that must be executed on all three rovers (e.g. map synchronization and driving), a separate task instance is created for each rover; constraints for that task are tied to the specific rover's state.
The state representation is identical for all rovers. Since the base station has no battery and doesn't drive, its state representation does not include battery SOC, and no driving tasks are created for it. The task network references a leader ID for tasks that must be executed on the leader (namely, team planning); but it does not have explicit knowledge of which agent is \gobble{currently} assigned as the leader.

For exploration, driving tasks for the participating rovers are independent of each other. If one rover is unable to drive for any reason, this does not impact the ability to schedule and execute driving for the other rovers. The distributed measurement experiment, however, requires coordinating tasks across all participating rovers. To achieve this, task hierarchy was used, a construct where a parent task decomposes into subtasks assigned to different rovers. The parent task is defined with the constraints and impacts of all subtasks, to ensure that the parent is scheduled at a time that is valid for all rovers. Then, the subtasks are constrained to be scheduled at the exact start and end of the parent. A multi-agent constraint is used to enforce continuous coordination of the tasks during execution. At the start of a formation drive, the parent task changes an internal state to ``coordinated'', while each subtask requires this state to be ``coordinated''. During execution, if one of the rover drives fails, it changes the state to ``not coordinated''. This creates a multi-agent constraint conflict that triggers an abort of the drive tasks on the other rovers, as described in Section \ref{sec:architecture:abort}.

\paragraph{Managing agent participation} The CADRE concept of operations allows operators to manually designate which agents participate in a given experiment. By default, all agents are included; however, ground operators can exclude one or more agents in case of failures, or to support anomaly investigations. The task networks support these alternate configurations using task constraints that require the state of the assigned agent to report it as ``participating''. For exploration, because tasks execute independently, the participation constraint is implemented similar to any other constraint. For formation sensing, however, driving is only scheduled if the tasks for all \emph{participating} rovers can be scheduled. We implement this by creating alternate hierarchies, one for each of the possible subsets of the three rovers. Parents with larger subsets are assigned higher priorities, so that they will be tried first. A subset that contains a subtask for a non-participating rover will fail that constraint, causing the full subset to fail. However, if a subtask for a rover fails to schedule for any other reason, we do not want to schedule a formation drive using the remaining rovers. Instead, we want to wait for all participating rovers to be ready to drive. This means that, for example, the subset for driving with only Rover 2 and Rover 3 includes a constraint that requires Rover 1 to be non-participating.

\subsection{Pytasknet}

CADRE's declarative task model was created using a Python library called Pytasknet. This library includes the basic classes and functions for creating tasks, and greatly simplified the effort of creating CADRE's task networks. The Python objects and references abstract away many of the details of creating references in MEXEC's XML task network representation. Moreover, the Python language provides the full power of a procedural language to create these task networks. For CADRE operations, where many tasks are identical except for the designated rover, the ability to create tasks within functions and loops was essential. \iftasknetworks \revI{A full representation of CADRE's task networks is reported \ifextendedversion in Appendix \ref{apx:tasknets}.\else in the Extended Version \cite{RabideauRussinoEaAAMAS24EV}.\fi}\fi %

\section{Experimental results and V\&V}
\label{sec:testing}

Throughout the development and V\&V process, the \pse subsystem for CADRE was tested over multiple venues at increasing levels of fidelity, as shown in Table \ref{tab:testing-levels}.

\begin{table}[h]
\caption{Testing venues used for validation of CADRE \pse}
\label{tab:testing-levels}
\vspace{-0.5em}
{\small
\begin{tabular}{c|ccccccccc}  
& \rotatebox{90}{Scheduling} &
 \rotatebox{90}{Nominal exec.} &
 \rotatebox{90}{Off-nominal exec.} &
 \rotatebox{90}{FSW integration} &
 \rotatebox{90}{Actual driving} &
 \rotatebox{90}{Outdoor driving} &
 \rotatebox{90}{Full sensor suite} &
 \rotatebox{90}{Specialized HW} &
 \rotatebox{90}{Flight HW} \\
 \hline
Batch planning & \checkmark & & & & & & & & \\
ROS sim & \checkmark& \checkmark & \checkmark & & & & & &\\
Dragonfarm & \checkmark & \checkmark & \checkmark & \checkmark & & & & \checkmark &\\
Dev. Models & \checkmark & \checkmark & \checkmark & \checkmark & \checkmark & \checkmark & & \checkmark &\\
Flight Models              & \checkmark & \checkmark & \checkmark  & \checkmark & \checkmark &  & \checkmark & \checkmark  & \checkmark \\
 \hline
\end{tabular}
}
\end{table}

Initial evaluation of the task networks and the scheduling implementation were performed using MEXEC batch planning, a one-shot process that produces a schedule based on a task network and a set of initial state values. To test the interplay between planning and execution, we conducted simulations first in a ROS-based \cite{quigley2009ros} simulator, with stand-ins for all system behaviors that have an impact on the planned tasks and the states they interact with. \revI{The simulator emulates task execution by idling for an appropriate duration, then returning the same fault/success codes used by FSW, and updating states based on the tasks' impact models. Individual tasks can return early, return late, or report failure, based on operator input.} For the ROS testing campaign, we developed a test matrix, Table \ref{table:1}, that crossed all of the tasks with all of the possible execution behaviors of a task. Each execution behavior (e.g. starting late, failing) can have a different impact on the re-planning and execution of subsequent tasks. In each relevant case, we verified that the proper response was taken by both the Agent Controller and the Strategic Planner. The ability to quickly and easily perform these tests was key to the development of the task network, and allowed us to identify multiple bugs where a task definition \gobble{worked for scheduling but }exhibited undesirable behavior during execution or re-scheduling\gobble{, when the state was different from the predicted state}. %

\begin{table}[h]
\centering
\caption{Testing campaign with ROS simulations. \gobble{SS = SSDB sync, TP = Team Planning, FS = Formation Sensing, EX = Exploration, ST = Stop, SL = Sleep. Some low priority tests are still pending.} As the first task to be scheduled, SSDB sync is not impacted by previous tasks and not expected to start late. Similarly, the stop and sleep tasks are not expected to have variable duration or fail. }
\label{table:1}
{\small
\begin{tabular}{c | c  c  c c  c } 
 Task & Nominal & Starts late & Runs late & Ends early & Fails \\ [0.5ex] 
 \hline%
 SSDB Sync & \checkmark &  & \checkmark & \checkmark & \checkmark \\ 
Team Planning & \checkmark &  & \checkmark & \checkmark & \checkmark \\
Formation & \checkmark & \checkmark & \checkmark & \checkmark & \checkmark \\
 Exploration & \checkmark & \checkmark & \checkmark & \checkmark & \checkmark \\
 Stop & \checkmark & \checkmark &  &  &  \\
 Sleep & \checkmark & \checkmark &  &  &  \\ [1ex] 
 \hline
\end{tabular}
}

\end{table}

Higher-fidelity simulations were then run on the \emph{Dragonfarm} testbed, which consists of a set of the same ModalAI VOXL system-on-a-chip modules used on the rovers, connected via Ethernet to provide a platform that can execute the full CADRE flight software deployment \revI{(which uses the F' framework \cite{bocchino2018fprime})}, with the same computing \gobble{(but not networking or mobility)} resources as flight hardware. In this context, we were able to perform extensive end-to-end simulation of both the exploration and formation sensing experiments with a variety of leader and participant configurations and initial state conditions.

\begin{figure}[h]
\centering
\includegraphics[width=.50\columnwidth]{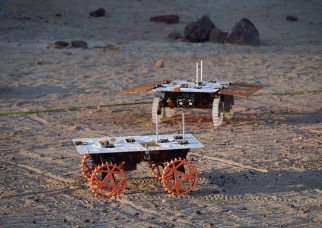}
\includegraphics[width=.475\columnwidth]{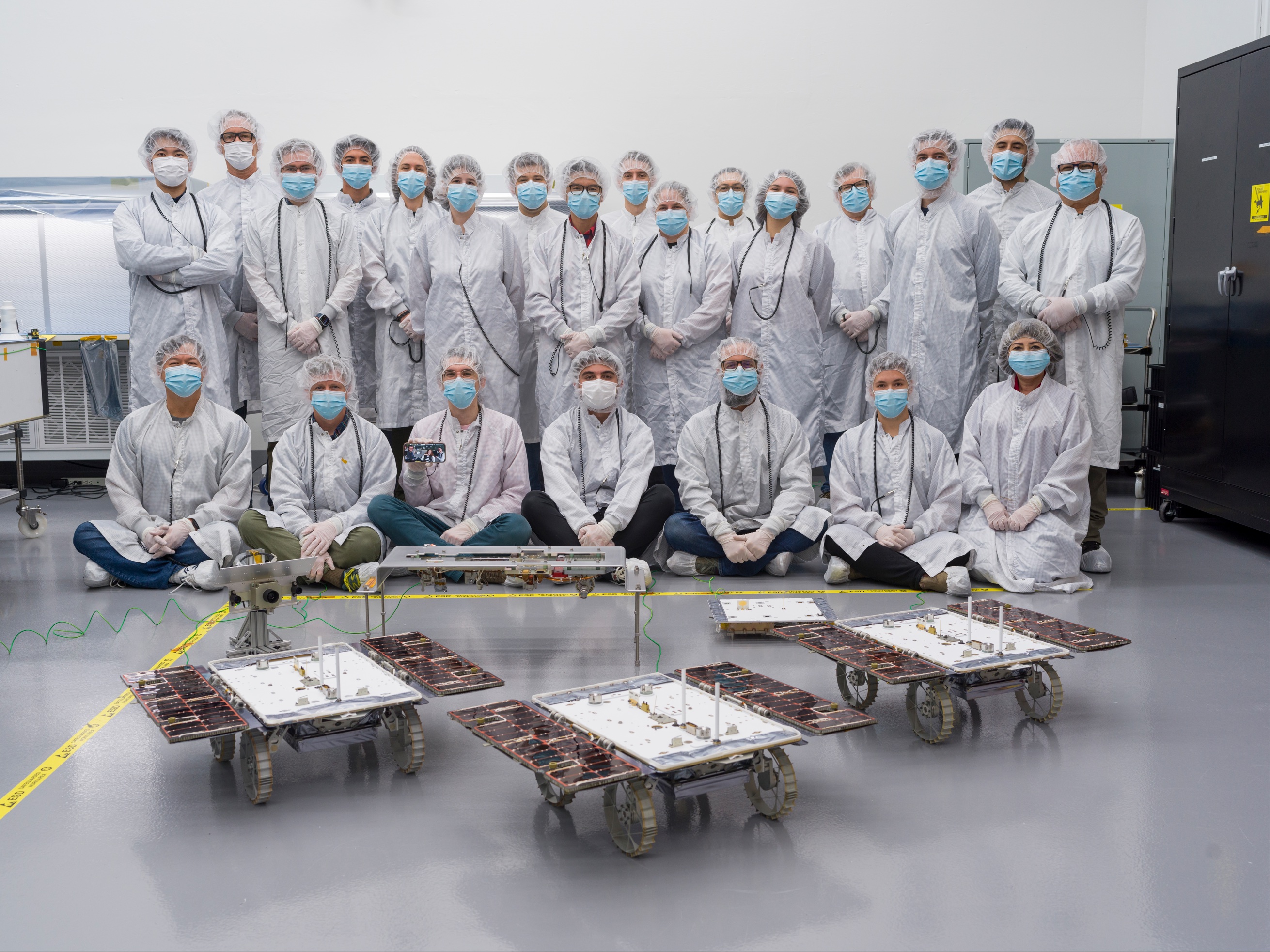}
\vspace{-1em}
\caption{CADRE's development models (DMs) in the Mars Yard (left) and flight models (FMs) in the cleanroom (right).}
\label{fig:dms-and-fms}
\end{figure}

Extensive testing was also carried out on mobility platforms -- specifically, Development Models (DM) and Flight Models (FM), shown in Figure \ref{fig:dms-and-fms}. The FMs are the hardware that will fly to Reiner Gamma; the DMs are reduced-fidelity hardware that present the same computing and actuation as the FMs, but a reduced sensor suite, and are suitable for outdoor testing on Earth. 

This multi-venue approach offered a comprehensive view of system performance. 
Testing on the FM rovers, despite their stringent operational constraints (e.g., being confined to the cleanroom environment, and subjected to rigorous safe handling guidelines), provided critical insights. The FM units were equipped with temperature and state-of-charge sensors \gobble{that are }not available on DMs, providing a complete picture of system performance and yielding high-fidelity data for testing the PS\&E system under flight-like conditions. In contrast, the DM rovers were not restricted to cleanroom conditions and were driven extensively in the Mars Yard outdoor testing venue, %
 which offered more freedom and terrain realism and allowed testing interactions with the Guidance, Navigation, and Control (GNC) subsystem. Balancing the strengths of both DM and FM testing was essential in ensuring that the autonomy system could handle both realistic driving tasks, and noisy sensor measurements.

Key tests on the DM hardware focused on running exploration and formation sensing experiments with flight-like wake-sleep cycles\gobble{, exercising as much of the stack as possible}. In these tests, the rovers autonomously woke up in autonomy mode, configured their sensors and components, and loaded the experiment task network into the strategic planner. The \pse stack handled all task planning, monitored task constraints during execution, and replanned as needed --- adjusting durations or replanning tasks without operator assistance or intervention.

Exploration testing was conducted across regions ranging from 6x6m to 22x13m. For distributed sensing, goal situated 6m to 20m from the agents' initial locations were tested. We verified that GNC could trigger replanning when the formation trajectory \revI{tube} couldn’t be followed, and that multi-agent constraint checking performed as expecting by stopping all agents in response. %
We also verified the ability to store and access state information across wake-sleep cycles, and demonstrated the ability to change the elected leader during different stages of the experiments. Testing under varying terrain configurations --- from sparse to cluttered rock and crater fields, and under the harsh shadows of night-time testing --- further validated the system’s robustness.

Key testing on the FM hardware involved Autonomy Day-in-the-Life experiments, where the stationary FM rovers performed tests under FPGA-driven wake/sleep operations. These tests ran with continuous cycles for over 9 hours, transitioning between autonomy, nominal, and safe modes. With real (albeit cleanroom-based) thermal measurements and battery state-of-charge calculations, we verified the planner's ability to generate appropriate plans based on thermal and power constraints. The system responded by either gracefully stopping experiments if constraints were violated, or changing task durations when limits were approaching. Additionally, the tests validated correct responses to off-nominal cases, such as sudden drops in battery state or temperature spikes, as well as agents entering safe mode. The tests also confirmed successful planning and state synchronization across all four agents, unlike the DM hardware, which only sports three agents.
Overall, not only did this testing campaign give us confidence that the distributed planning, scheduling, and execution system can perform under the scenarios required by the CADRE mission, but also gave us key insights into how to operate autonomous and multi agent spacecraft.

\section{Conclusion}
\label{sec:conclusion}

In this paper, we describe the planning, scheduling, and execution architecture for CADRE, a NASA mission that will demonstrate multi-agent autonomy on the Moon in \revI{2025/2026}. Testing of the autonomy stack also provided unique insight into verification and validation of autonomy, and operations of multi-agent systems. Successful completion of CADRE's mission will demonstrate the high technology readiness of multi-agent planning, scheduling, and execution in challenging planetary surface environments (in particular, on the Moon), enabling infusion in future science-driven planetary exploration missions.

\begin{acks}
This research was carried out at the Jet Propulsion Laboratory, California Institute of Technology, under a contract with the National Aeronautics and Space Administration (80NM0018D0004).
\end{acks}
\newpage

\bibliographystyle{ACM-Reference-Format} 
\bibliography{bibliography}


\begin{thebibliography}{32}


\ifx \showCODEN    \undefined \def \showCODEN     #1{\unskip}     \fi
\ifx \showDOI      \undefined \def \showDOI       #1{#1}\fi
\ifx \showISBNx    \undefined \def \showISBNx     #1{\unskip}     \fi
\ifx \showISBNxiii \undefined \def \showISBNxiii  #1{\unskip}     \fi
\ifx \showISSN     \undefined \def \showISSN      #1{\unskip}     \fi
\ifx \showLCCN     \undefined \def \showLCCN      #1{\unskip}     \fi
\ifx \shownote     \undefined \def \shownote      #1{#1}          \fi
\ifx \showarticletitle \undefined \def \showarticletitle #1{#1}   \fi
\ifx \showURL      \undefined \def \showURL       {\relax}        \fi
\providecommand\bibfield[2]{#2}
\providecommand\bibinfo[2]{#2}
\providecommand\natexlab[1]{#1}
\providecommand\showeprint[2][]{arXiv:#2}

\bibitem[\protect\citeauthoryear{Adams and Frank}{Adams and Frank}{2024}]%
        {adams2024distributed}
\bibfield{author}{\bibinfo{person}{Caleb Adams} {and} \bibinfo{person}{Jeremy
  Frank}.} \bibinfo{year}{2024}\natexlab{}.
\newblock \bibinfo{title}{The Distributed Spacecraft Autonomy (DSA)}.
  (\bibinfo{year}{2024}).
\newblock
\newblock
\shownote{34th International Conference on Automated Planning and Scheduling
  (ICAPS) Summer School.}


\bibitem[\protect\citeauthoryear{Adams, Kempa, Iatauro, Frank, and
  Vaughan}{Adams et~al\mbox{.}}{2023}]%
        {adams2023overview}
\bibfield{author}{\bibinfo{person}{Caleb Adams}, \bibinfo{person}{Brian Kempa},
  \bibinfo{person}{Michael Iatauro}, \bibinfo{person}{Jeremy Frank}, {and}
  \bibinfo{person}{Walter Vaughan}.} \bibinfo{year}{2023}\natexlab{}.
\newblock \showarticletitle{An Overview of Distributed Spacecraft Autonomy at
  NASA Ames}. In \bibinfo{booktitle}{\emph{Proceedings of the Small Satellite
  Conference}}. Utah State University.
\newblock


\bibitem[\protect\citeauthoryear{Albee, Bhamidipati, Rossi, Vander~Hook, and
  de~la Croix}{Albee et~al\mbox{.}}{2024}]%
        {AlbeeBhamidipatiEa2024}
\bibfield{author}{\bibinfo{person}{Keenan Albee}, \bibinfo{person}{Sriramya
  Bhamidipati}, \bibinfo{person}{Federico Rossi}, \bibinfo{person}{Joshua
  Vander~Hook}, {and} \bibinfo{person}{Jean-Pierre de~la Croix}.}
  \bibinfo{year}{2024}\natexlab{}.
\newblock \showarticletitle{Lunar Leader: Persistent, Optimal Leader Election
  for Multi-Agent Exploration Teams}. In \bibinfo{booktitle}{\emph{{Int.
  Workshop on Autonomous Agents and Multi-Agent Systems for Space Applications
  (MASSpace)}}}. \bibinfo{address}{Auckland, NZ}.
\newblock


\bibitem[\protect\citeauthoryear{Bocchino, Canham, Watney, Reder, and
  Levison}{Bocchino et~al\mbox{.}}{2018}]%
        {bocchino2018fprime}
\bibfield{author}{\bibinfo{person}{Robert Bocchino}, \bibinfo{person}{Timothy
  Canham}, \bibinfo{person}{Garth Watney}, \bibinfo{person}{Leonard Reder},
  {and} \bibinfo{person}{Jeffrey Levison}.} \bibinfo{year}{2018}\natexlab{}.
\newblock \showarticletitle{F Prime: an open-source framework for small-scale
  flight software systems}. In \bibinfo{booktitle}{\emph{Small Satellite
  Conference}}. \bibinfo{address}{Logan, UT}.
\newblock


\bibitem[\protect\citeauthoryear{Cellucci, Cramer, and Frank}{Cellucci
  et~al\mbox{.}}{2020}]%
        {cellucci2020distributed}
\bibfield{author}{\bibinfo{person}{Daniel Cellucci}, \bibinfo{person}{Nick~B
  Cramer}, {and} \bibinfo{person}{Jeremy~D Frank}.}
  \bibinfo{year}{2020}\natexlab{}.
\newblock \showarticletitle{Distributed spacecraft autonomy}. In
  \bibinfo{booktitle}{\emph{ASCEND 2020}}. \bibinfo{pages}{4232}.
\newblock


\bibitem[\protect\citeauthoryear{Chien, Sherwood, Tran, Cichy, Rabideau,
  Castano, Davies, Mandl, Frye, Trout, Shulman, and Boyer}{Chien
  et~al\mbox{.}}{2005}]%
        {chien2005eo1}
\bibfield{author}{\bibinfo{person}{S. Chien}, \bibinfo{person}{R. Sherwood},
  \bibinfo{person}{D. Tran}, \bibinfo{person}{B. Cichy}, \bibinfo{person}{G.
  Rabideau}, \bibinfo{person}{R. Castano}, \bibinfo{person}{A. Davies},
  \bibinfo{person}{D. Mandl}, \bibinfo{person}{S. Frye}, \bibinfo{person}{B.
  Trout}, \bibinfo{person}{S. Shulman}, {and} \bibinfo{person}{D. Boyer}.}
  \bibinfo{year}{2005}\natexlab{}.
\newblock \bibinfo{title}{Using Autonomy Flight Software to Improve Science
  Return on Earth Observing One}.
\newblock , \bibinfo{numpages}{196--216}~pages.
\newblock


\bibitem[\protect\citeauthoryear{Choi, Brunet, and How}{Choi
  et~al\mbox{.}}{2009}]%
        {choi2009consensus}
\bibfield{author}{\bibinfo{person}{Han-Lim Choi}, \bibinfo{person}{Luc Brunet},
  {and} \bibinfo{person}{Jonathan~P How}.} \bibinfo{year}{2009}\natexlab{}.
\newblock \showarticletitle{Consensus-based decentralized auctions for robust
  task allocation}.
\newblock \bibinfo{journal}{\emph{IEEE transactions on robotics}}
  \bibinfo{volume}{25}, \bibinfo{number}{4} (\bibinfo{year}{2009}),
  \bibinfo{pages}{912--926}.
\newblock


\bibitem[\protect\citeauthoryear{de~la Croix, Rossi, Brockers, Aguilar, Albee,
  Boroson, Cauligi, Delaune, Hewitt, Kogan, Lim, Morrell, Nakka, Nguyen,
  Proenca, Rabideau, Russino, da~Silva, Zohar, and Comandur}{de~la Croix
  et~al\mbox{.}}{2024}]%
        {DeLaCroixRossiEa2024}
\bibfield{author}{\bibinfo{person}{Jean-Pierre de~la Croix},
  \bibinfo{person}{Federico Rossi}, \bibinfo{person}{Roland Brockers},
  \bibinfo{person}{Dustin Aguilar}, \bibinfo{person}{Keenan Albee},
  \bibinfo{person}{Elizabeth Boroson}, \bibinfo{person}{Abhishek Cauligi},
  \bibinfo{person}{Jeff Delaune}, \bibinfo{person}{Robert Hewitt},
  \bibinfo{person}{Dima Kogan}, \bibinfo{person}{Grace Lim},
  \bibinfo{person}{Benjamin Morrell}, \bibinfo{person}{Yashwanth Nakka},
  \bibinfo{person}{Viet Nguyen}, \bibinfo{person}{Pedro Proenca},
  \bibinfo{person}{Gregg Rabideau}, \bibinfo{person}{Joseph Russino},
  \bibinfo{person}{Maira~Saboia da Silva}, \bibinfo{person}{Guy Zohar}, {and}
  \bibinfo{person}{Subha Comandur}.} \bibinfo{year}{2024}\natexlab{}.
\newblock \showarticletitle{Multi-Agent Autonomy for Space Exploration on the
  CADRE Lunar Technology Demonstration}. In \bibinfo{booktitle}{\emph{{IEEE
  Aerospace Conference}}}. \bibinfo{address}{Big Sky, MT},
  \bibinfo{pages}{1--14}.
\newblock
\showISSN{1095-323X}
\urldef\tempurl%
\url{https://doi.org/10.1109/AERO58975.2024.10521425}
\showDOI{\tempurl}


\bibitem[\protect\citeauthoryear{Gaines, Doran, Justice, Rabideau, Schaffer,
  Verma, Wagstaff, Vasavada, Huffman, Anderson, Mackey, and Estlin}{Gaines
  et~al\mbox{.}}{2017}]%
        {gaines-doran-justice-et-al-IWPSS-2017}
\bibfield{author}{\bibinfo{person}{Daniel Gaines}, \bibinfo{person}{Gary
  Doran}, \bibinfo{person}{Heather Justice}, \bibinfo{person}{Gregg Rabideau},
  \bibinfo{person}{Steve Schaffer}, \bibinfo{person}{Vandana Verma},
  \bibinfo{person}{Kiri Wagstaff}, \bibinfo{person}{Ashwin Vasavada},
  \bibinfo{person}{William Huffman}, \bibinfo{person}{Robert Anderson},
  \bibinfo{person}{Ryan Mackey}, {and} \bibinfo{person}{Tara Estlin}.}
  \bibinfo{year}{2017}\natexlab{}.
\newblock \showarticletitle{A Case Study of Productivity Challenges in Mars
  Science Laboratory Operations}. In \bibinfo{booktitle}{\emph{International
  Workshop on Planning and Scheduling for Space (IWPSS 2017)}}.
  \bibinfo{address}{Pittsburgh, PA}.
\newblock


\bibitem[\protect\citeauthoryear{Gaines, Doran, Justice, Rabideau, Schaffer,
  Verma, Wagstaff, Vasavada, Huffman, Anderson, Mackey, and Estlin}{Gaines
  et~al\mbox{.}}{2016}]%
        {gaines-doran-justice-et-al-2016}
\bibfield{author}{\bibinfo{person}{D. Gaines}, \bibinfo{person}{G. Doran},
  \bibinfo{person}{H. Justice}, \bibinfo{person}{G. Rabideau},
  \bibinfo{person}{S. Schaffer}, \bibinfo{person}{V. Verma},
  \bibinfo{person}{K. Wagstaff}, \bibinfo{person}{V. Vasavada},
  \bibinfo{person}{W. Huffman}, \bibinfo{person}{R. Anderson},
  \bibinfo{person}{R. Mackey}, {and} \bibinfo{person}{T. Estlin}.}
  \bibinfo{year}{2016}\natexlab{}.
\newblock \bibinfo{booktitle}{\emph{Productivity challenges for Mars rover
  operations: A case study of Mars Science Laboratory operations.}}
\newblock \bibinfo{type}{{T}echnical {R}eport} D-97908.
  \bibinfo{institution}{Jet Propulsion Laboratory}.
\newblock


\bibitem[\protect\citeauthoryear{Gallager, Humblet, and Spira}{Gallager
  et~al\mbox{.}}{1983}]%
        {gallager1983distributed}
\bibfield{author}{\bibinfo{person}{Robert~G. Gallager},
  \bibinfo{person}{Pierre~A. Humblet}, {and} \bibinfo{person}{Philip~M.
  Spira}.} \bibinfo{year}{1983}\natexlab{}.
\newblock \showarticletitle{A distributed algorithm for minimum-weight spanning
  trees}.
\newblock \bibinfo{journal}{\emph{ACM Transactions on Programming Languages and
  systems (TOPLAS)}} \bibinfo{volume}{5}, \bibinfo{number}{1}
  (\bibinfo{year}{1983}), \bibinfo{pages}{66--77}.
\newblock


\bibitem[\protect\citeauthoryear{Gilbert and Lynch}{Gilbert and Lynch}{2002}]%
        {gilbert2002brewer}
\bibfield{author}{\bibinfo{person}{Seth Gilbert} {and} \bibinfo{person}{Nancy
  Lynch}.} \bibinfo{year}{2002}\natexlab{}.
\newblock \showarticletitle{Brewer's conjecture and the feasibility of
  consistent, available, partition-tolerant web services}.
\newblock \bibinfo{journal}{\emph{Acm Sigact News}} \bibinfo{volume}{33},
  \bibinfo{number}{2} (\bibinfo{year}{2002}), \bibinfo{pages}{51--59}.
\newblock


\bibitem[\protect\citeauthoryear{Gray}{Gray}{1978}]%
        {Gray1978}
\bibfield{author}{\bibinfo{person}{J.~N. Gray}.}
  \bibinfo{year}{1978}\natexlab{}.
\newblock \bibinfo{booktitle}{\emph{Notes on data base operating systems}}.
\newblock \bibinfo{publisher}{Springer Berlin Heidelberg},
  \bibinfo{address}{Berlin, Heidelberg}, \bibinfo{pages}{393--481}.
\newblock
\showISBNx{978-3-540-35880-0}
\urldef\tempurl%
\url{https://doi.org/10.1007/3-540-08755-9_9}
\showDOI{\tempurl}


\bibitem[\protect\citeauthoryear{Haberle, G{\'o}mez-Elvira, de~la
  Torre~Ju{\'a}rez, Harri, Hollingsworth, Kahanp{\"a}{\"a}, Kahre, Lemmon,
  Mart{\'\i}n-Torres, Mischna, et~al\mbox{.}}{Haberle et~al\mbox{.}}{2014}]%
        {haberle2014preliminary}
\bibfield{author}{\bibinfo{person}{RM Haberle}, \bibinfo{person}{Javier
  G{\'o}mez-Elvira}, \bibinfo{person}{M de~la Torre~Ju{\'a}rez},
  \bibinfo{person}{A-M Harri}, \bibinfo{person}{JL Hollingsworth},
  \bibinfo{person}{Henrik Kahanp{\"a}{\"a}}, \bibinfo{person}{MA Kahre},
  \bibinfo{person}{M Lemmon}, \bibinfo{person}{FJ Mart{\'\i}n-Torres},
  \bibinfo{person}{M Mischna}, {et~al\mbox{.}}}
  \bibinfo{year}{2014}\natexlab{}.
\newblock \showarticletitle{Preliminary interpretation of the {REMS} pressure
  data from the first 100 sols of the {MSL} mission}.
\newblock \bibinfo{journal}{\emph{Journal of Geophysical Research: Planets}}
  \bibinfo{volume}{119}, \bibinfo{number}{3} (\bibinfo{year}{2014}),
  \bibinfo{pages}{440--453}.
\newblock


\bibitem[\protect\citeauthoryear{Karaman and Frazzoli}{Karaman and
  Frazzoli}{2011}]%
        {karaman2011sampling}
\bibfield{author}{\bibinfo{person}{Sertac Karaman} {and}
  \bibinfo{person}{Emilio Frazzoli}.} \bibinfo{year}{2011}\natexlab{}.
\newblock \showarticletitle{Sampling-based algorithms for optimal motion
  planning}.
\newblock \bibinfo{journal}{\emph{The International Journal of Robotics
  Research}} \bibinfo{volume}{30}, \bibinfo{number}{7} (\bibinfo{year}{2011}),
  \bibinfo{pages}{846--894}.
\newblock


\bibitem[\protect\citeauthoryear{Mankins}{Mankins}{1995}]%
        {mankins1995technology}
\bibfield{author}{\bibinfo{person}{John~C Mankins}.}
  \bibinfo{year}{1995}\natexlab{}.
\newblock \bibinfo{booktitle}{\emph{Technology readiness levels}}.
\newblock \bibinfo{type}{{T}echnical {R}eport}. \bibinfo{institution}{Advanced
  Concepts Office, Office of Space Access and Technology, NASA}.
\newblock


\bibitem[\protect\citeauthoryear{Nayak, Rossi, Lim, Otte, and de~la
  Croix}{Nayak et~al\mbox{.}}{2024}]%
        {nayak2024exploration}
\bibfield{author}{\bibinfo{person}{Sharan Nayak}, \bibinfo{person}{Federico
  Rossi}, \bibinfo{person}{Grace Lim}, \bibinfo{person}{Michael~W. Otte}, {and}
  \bibinfo{person}{Jean-Pierre de~la Croix}.} \bibinfo{year}{2024}\natexlab{}.
\newblock \bibinfo{title}{Multi-Robot Exploration for the CADRE Mission}.
  (\bibinfo{year}{2024}).
\newblock
\newblock
\shownote{Under review.}


\bibitem[\protect\citeauthoryear{{NetSAG}}{{NetSAG}}{2010}]%
        {netsag2010}
\bibfield{author}{\bibinfo{person}{{NetSAG}}.} \bibinfo{year}{2010}\natexlab{}.
\newblock \bibinfo{booktitle}{\emph{Mars Network Science Analysis Group
  ({NetSAG}) Final Report}}.
\newblock \bibinfo{type}{{T}echnical {R}eport}. \bibinfo{institution}{{MEPAG}}.
\newblock
\urldef\tempurl%
\url{https://mepag.jpl.nasa.gov/meeting/2010-03/NetSAG_MEPAG_Final_v2.pdf}
\showURL{%
\tempurl}


\bibitem[\protect\citeauthoryear{Parjan and Chien}{Parjan and Chien}{2023}]%
        {parjan-jais2023-mas}
\bibfield{author}{\bibinfo{person}{Shreya Parjan} {and} \bibinfo{person}{Steve
  Chien}.} \bibinfo{year}{2023}\natexlab{}.
\newblock \showarticletitle{Decentralized Observation Allocation for a
  Large-Scale Constellation}.
\newblock \bibinfo{journal}{\emph{Journal of Aerospace Information Systems
  (JAIS)}} \bibinfo{volume}{20}, \bibinfo{number}{8} (\bibinfo{date}{August}
  \bibinfo{year}{2023}), \bibinfo{pages}{447--461}.
\newblock
\urldef\tempurl%
\url{https://doi.org/10.2514/1.I011215}
\showDOI{\tempurl}


\bibitem[\protect\citeauthoryear{Quigley, Conley, Gerkey, Faust, Foote, Leibs,
  Wheeler, Ng, et~al\mbox{.}}{Quigley et~al\mbox{.}}{2009}]%
        {quigley2009ros}
\bibfield{author}{\bibinfo{person}{Morgan Quigley}, \bibinfo{person}{Ken
  Conley}, \bibinfo{person}{Brian Gerkey}, \bibinfo{person}{Josh Faust},
  \bibinfo{person}{Tully Foote}, \bibinfo{person}{Jeremy Leibs},
  \bibinfo{person}{Rob Wheeler}, \bibinfo{person}{Andrew~Y Ng},
  {et~al\mbox{.}}} \bibinfo{year}{2009}\natexlab{}.
\newblock \showarticletitle{ROS: an open-source Robot Operating System}. In
  \bibinfo{booktitle}{\emph{ICRA workshop on open source software}},
  Vol.~\bibinfo{volume}{3}. \bibinfo{address}{Kobe, Japan}, \bibinfo{pages}{5}.
\newblock


\bibitem[\protect\citeauthoryear{Rossi, Bandyopadhyay, Wolf, and Pavone}{Rossi
  et~al\mbox{.}}{2021}]%
        {RossiBandyopadhyayEtAl2021}
\bibfield{author}{\bibinfo{person}{Federico Rossi}, \bibinfo{person}{Saptarshi
  Bandyopadhyay}, \bibinfo{person}{Michael~T. Wolf}, {and}
  \bibinfo{person}{Marco Pavone}.} \bibinfo{year}{2021}\natexlab{}.
\newblock \bibinfo{title}{Multi-Agent Algorithms for Collective Behavior - A
  structural and application-focused atlas}.  (\bibinfo{year}{2021}).
\newblock
\urldef\tempurl%
\url{https://arxiv.org/abs/2103.11067}
\showURL{%
\tempurl}


\bibitem[\protect\citeauthoryear{Rossi, Saboia, Krishnamoorthy, and
  Vander~Hook}{Rossi et~al\mbox{.}}{2023}]%
        {RossiSaboiaEtAl2023}
\bibfield{author}{\bibinfo{person}{Federico Rossi}, \bibinfo{person}{Ma\'ira
  Saboia}, \bibinfo{person}{Siddharth Krishnamoorthy}, {and}
  \bibinfo{person}{Joshua Vander~Hook}.} \bibinfo{year}{2023}\natexlab{}.
\newblock \showarticletitle{Proximal Exploration of Venus Volcanism with Teams
  of Autonomous Buoyancy-Controlled Balloons}.
\newblock \bibinfo{journal}{\emph{{Acta Astronautica}}}  \bibinfo{volume}{208}
  (\bibinfo{year}{2023}), \bibinfo{pages}{389--406}.
\newblock
\urldef\tempurl%
\url{https://doi.org/10.1016/j.actaastro.2023.03.003}
\showDOI{\tempurl}


\bibitem[\protect\citeauthoryear{Saboia, Rossi, Nguyen, Lim, Aguilar, and de~la
  Croix}{Saboia et~al\mbox{.}}{2024}]%
        {SaboiaRossiEa2024}
\bibfield{author}{\bibinfo{person}{Maira Saboia}, \bibinfo{person}{Federico
  Rossi}, \bibinfo{person}{Viet Nguyen}, \bibinfo{person}{Grace Lim},
  \bibinfo{person}{Dustin Aguilar}, {and} \bibinfo{person}{Jean-Pierre de~la
  Croix}.} \bibinfo{year}{2024}\natexlab{}.
\newblock \showarticletitle{CADRE MoonDB: Distributed Database for Multi-Robot
  Information-Sharing and Map-Merging for Lunar Exploration}. In
  \bibinfo{booktitle}{\emph{{Int. Workshop on Autonomous Agents and Multi-Agent
  Systems for Space Applications (MASSpace)}}}. \bibinfo{address}{Auckland,
  NZ}.
\newblock


\bibitem[\protect\citeauthoryear{Sellmaier, Uhlig, and Schmidhuber}{Sellmaier
  et~al\mbox{.}}{2022}]%
        {sellmaier2022spacecraft}
\bibfield{author}{\bibinfo{person}{Florian Sellmaier}, \bibinfo{person}{Thomas
  Uhlig}, {and} \bibinfo{person}{Michael Schmidhuber}.}
  \bibinfo{year}{2022}\natexlab{}.
\newblock \bibinfo{booktitle}{\emph{Spacecraft Operations}}.
\newblock \bibinfo{publisher}{Springer}.
\newblock


\bibitem[\protect\citeauthoryear{Troesch, Mirza, Hughes, Rothstein-Dowden,
  Bocchino, Donner, Feather, Smith, Fesq, Barker, and Campuzano}{Troesch
  et~al\mbox{.}}{2020}]%
        {troesch_mexec_asteria_intex2020}
\bibfield{author}{\bibinfo{person}{M. Troesch}, \bibinfo{person}{F. Mirza},
  \bibinfo{person}{K. Hughes}, \bibinfo{person}{A. Rothstein-Dowden},
  \bibinfo{person}{R. Bocchino}, \bibinfo{person}{A. Donner},
  \bibinfo{person}{M. Feather}, \bibinfo{person}{B. Smith}, \bibinfo{person}{L.
  Fesq}, \bibinfo{person}{B. Barker}, {and} \bibinfo{person}{B. Campuzano}.}
  \bibinfo{year}{2020}\natexlab{}.
\newblock \showarticletitle{MEXEC: An Onboard Integrated Planning and Execution
  Approach for Spacecraft Commanding}. In \bibinfo{booktitle}{\emph{Workshop on
  Integrated Execution (IntEx) / Goal Reasoning (GR), International Conference
  on Automated Planning and Scheduling (ICAPS IntEx/GP 2020)}}.
\newblock
\urldef\tempurl%
\url{https://ai.jpl.nasa.gov/public/papers/IntEx-2020-MEXEC.pdf}
\showURL{%
\tempurl}
\newblock
\shownote{Also presented at International Symposium on Artificial Intelligence,
  Robotics, and Automation for Space (i-SAIRAS 2020) and appears as an
  abstract.}


\bibitem[\protect\citeauthoryear{Troesch, Mirza, Rabideau, and Chien}{Troesch
  et~al\mbox{.}}{2019}]%
        {troesch_iwpss2019_robustmapping}
\bibfield{author}{\bibinfo{person}{M. Troesch}, \bibinfo{person}{F. Mirza},
  \bibinfo{person}{G. Rabideau}, {and} \bibinfo{person}{S. Chien}.}
  \bibinfo{year}{2019}\natexlab{}.
\newblock \showarticletitle{Onboard re-planning for robust mapping using
  pre-compiled backup observations}. In \bibinfo{booktitle}{\emph{11th
  International Workshop on Planning and Scheduling for Space (IWPSS)}}.
  \bibinfo{address}{Berkeley, California, USA}, \bibinfo{pages}{168--175}.
\newblock


\bibitem[\protect\citeauthoryear{Vance, Kedar, Panning, St{\"{a}}hler, Bills,
  Lorenz, Huang, Pike, Castillo, Lognonn{\'{e}}, Tsai, and Rhoden}{Vance
  et~al\mbox{.}}{2018}]%
        {Vance}
\bibfield{author}{\bibinfo{person}{Steven~D. Vance}, \bibinfo{person}{Sharon
  Kedar}, \bibinfo{person}{Mark~P. Panning}, \bibinfo{person}{Simon~C.
  St{\"{a}}hler}, \bibinfo{person}{Bruce~G. Bills}, \bibinfo{person}{Ralph~D.
  Lorenz}, \bibinfo{person}{Hsin-Hua Huang}, \bibinfo{person}{W.T. Pike},
  \bibinfo{person}{Julie~C. Castillo}, \bibinfo{person}{Philippe
  Lognonn{\'{e}}}, \bibinfo{person}{Victor~C. Tsai}, {and}
  \bibinfo{person}{Alyssa~R. Rhoden}.} \bibinfo{year}{2018}\natexlab{}.
\newblock \showarticletitle{{Vital Signs: Seismology of Icy Ocean Worlds}}.
\newblock \bibinfo{journal}{\emph{Astrobiology}} \bibinfo{volume}{18},
  \bibinfo{number}{1} (\bibinfo{date}{Jan.} \bibinfo{year}{2018}),
  \bibinfo{pages}{37--53}.
\newblock
\showISSN{1531-1074}
\urldef\tempurl%
\url{https://doi.org/10.1089/ast.2016.1612}
\showDOI{\tempurl}


\bibitem[\protect\citeauthoryear{Verma, Maimone, Gaines, Francis, Estlin, Kuhn,
  Rabideau, Chien, McHenry, Graser, et~al\mbox{.}}{Verma et~al\mbox{.}}{2023}]%
        {verma2023autonomous}
\bibfield{author}{\bibinfo{person}{Vandi Verma}, \bibinfo{person}{Mark~W
  Maimone}, \bibinfo{person}{Daniel~M Gaines}, \bibinfo{person}{Raymond
  Francis}, \bibinfo{person}{Tara~A Estlin}, \bibinfo{person}{Stephen~R Kuhn},
  \bibinfo{person}{Gregg~R Rabideau}, \bibinfo{person}{Steve~A Chien},
  \bibinfo{person}{Michael~M McHenry}, \bibinfo{person}{Evan~J Graser},
  {et~al\mbox{.}}} \bibinfo{year}{2023}\natexlab{}.
\newblock \showarticletitle{Autonomous robotics is driving Perseverance
  rover’s progress on Mars}.
\newblock \bibinfo{journal}{\emph{Science Robotics}} \bibinfo{volume}{8},
  \bibinfo{number}{80} (\bibinfo{year}{2023}).
\newblock


\bibitem[\protect\citeauthoryear{Werfel, Petersen, and Nagpal}{Werfel
  et~al\mbox{.}}{2014}]%
        {Werfel2014termite}
\bibfield{author}{\bibinfo{person}{Justin Werfel}, \bibinfo{person}{Kirstin
  Petersen}, {and} \bibinfo{person}{Radhika Nagpal}.}
  \bibinfo{year}{2014}\natexlab{}.
\newblock \showarticletitle{Designing Collective Behavior in a Termite-Inspired
  Robot Construction Team}.
\newblock \bibinfo{journal}{\emph{Science}} \bibinfo{volume}{343},
  \bibinfo{number}{6172} (\bibinfo{year}{2014}), \bibinfo{pages}{754--758}.
\newblock
\urldef\tempurl%
\url{https://doi.org/10.1126/science.1245842}
\showDOI{\tempurl}


\bibitem[\protect\citeauthoryear{Wolf, Rahmani, de~la Croix, Woodward,
  Vander~Hook, Brown, Schaffer, Lim, Bailey, Tepsuporn, et~al\mbox{.}}{Wolf
  et~al\mbox{.}}{2017}]%
        {wolf2017caracas}
\bibfield{author}{\bibinfo{person}{Michael~T Wolf}, \bibinfo{person}{Amir
  Rahmani}, \bibinfo{person}{Jean-Pierre de~la Croix}, \bibinfo{person}{Gail
  Woodward}, \bibinfo{person}{Joshua Vander~Hook}, \bibinfo{person}{David
  Brown}, \bibinfo{person}{Steve Schaffer}, \bibinfo{person}{Christopher Lim},
  \bibinfo{person}{Philip Bailey}, \bibinfo{person}{Scott Tepsuporn},
  {et~al\mbox{.}}} \bibinfo{year}{2017}\natexlab{}.
\newblock \showarticletitle{CARACaS multi-agent maritime autonomy for unmanned
  surface vehicles in the Swarm II harbor patrol demonstration}. In
  \bibinfo{booktitle}{\emph{Unmanned systems technology XIX}},
  Vol.~\bibinfo{volume}{10195}. SPIE, \bibinfo{pages}{218--228}.
\newblock


\bibitem[\protect\citeauthoryear{Yamauchi}{Yamauchi}{1997}]%
        {yamauchi1997frontier}
\bibfield{author}{\bibinfo{person}{Brian Yamauchi}.}
  \bibinfo{year}{1997}\natexlab{}.
\newblock \showarticletitle{A frontier-based approach for autonomous
  exploration}. In \bibinfo{booktitle}{\emph{Proceedings 1997 IEEE
  International Symposium on Computational Intelligence in Robotics and
  Automation CIRA'97.'Towards New Computational Principles for Robotics and
  Automation'}}. IEEE, \bibinfo{pages}{146--151}.
\newblock


\bibitem[\protect\citeauthoryear{Zilberstein, Rao, Salis, and
  Chien}{Zilberstein et~al\mbox{.}}{2024}]%
        {zilberstein-icaps-2024}
\bibfield{author}{\bibinfo{person}{Itai Zilberstein}, \bibinfo{person}{Ananya
  Rao}, \bibinfo{person}{Matthew Salis}, {and} \bibinfo{person}{Steve Chien}.}
  \bibinfo{year}{2024}\natexlab{}.
\newblock \showarticletitle{Decentralized, Decomposition-Based Observation
  Scheduling for a Large-Scale Satellite Constellation}. In
  \bibinfo{booktitle}{\emph{International Conference on Automated Planning and
  Scheduling}}. \bibinfo{address}{Banff, Canada}.
\newblock
\urldef\tempurl%
\url{https://ai.jpl.nasa.gov/public/documents/papers/Zilberstein-ICAPS-2024.pdf}
\showURL{%
\tempurl}


\end{thebibliography}
\balance

\ifextendedversion
\newpage
\onecolumn
\begin{appendix}

\section{Scheduling Algorithm}\label{apx:simpleplanner}

\begin{algorithm}[H]
\caption{MEXEC\_PLAN\_SEARCH\_DECOMP\_PRIORITY}\label{alg:decomp}
\begin{algorithmic}
    \For{the maximum number of tasks in the network}
        \For{the maximum number of scheduling iterations}
            \State consider next unscheduled task with a preferred start time in the scheduling window
            \If {there is a valid interval in which to schedule the task}
                \State schedule the task as close to its preferred start time in the valid interval
                \If {task has a detail command}
                    \State dispatch detail command
                \EndIf
                \If {task has a decomposition}
                    \State create subtasks from templates and add them to the unscheduled tasks list
                \EndIf
            \EndIf
        \EndFor
        \If {no tasks were scheduled}
            \State break
        \EndIf
    \EndFor
\end{algorithmic}
\end{algorithm}

\iftasknetworks
\section{Task networks}\label{apx:tasknets}
\subsection{Exploration}

\subsubsection{Task network}
{\scriptsize
\inputminted{python}{code/strategic-planner/tasknets/gen_explore_tasknet.py}
}

\subsubsection{Tasks}

{\scriptsize
\inputminted{python}{code/strategic-planner/tasknets/etp.py}
}

\subsection{Distributed Measurement}
\subsubsection{Task network}
{\scriptsize
\inputminted{python}{code/strategic-planner/tasknets/gen_formation_sense_tasknet.py}
}

\subsection{Common Tasks}
\subsubsection{Sleep task}
{\scriptsize
\inputminted{python}{code/strategic-planner/tasknets/sleep.py}
}
\subsubsection{Shared state DB synchronization task}

{\scriptsize
\inputminted{python}{code/strategic-planner/tasknets/ssdb.py}
}

\subsubsection{Stop task}
{\scriptsize
\inputminted{python}{code/strategic-planner/tasknets/stop.py}
}

\fi

\end{appendix}
\fi

\end{document}